\documentclass[letterpaper, 10 pt, conference]{ieeeconf}  

\IEEEoverridecommandlockouts                              

\usepackage{graphics} 
\usepackage{times} 
\usepackage{amsmath} 
\usepackage{amssymb}  
\usepackage{algpseudocode,algorithm,algorithmicx}

\usepackage{caption}
\usepackage{graphicx}
\usepackage{multirow}
\usepackage{bm}
\usepackage{booktabs}
\usepackage{array}
\usepackage{gensymb} 
\usepackage{subcaption}
\usepackage[keeplastbox]{flushend}

\newcolumntype{C}[1]{>{\centering}m{#1}}
\DeclareMathOperator*{\argmax}{argmax}

\title{\LARGE \bf
Coverage Path Planning using Path Primitive Sampling and Primitive Coverage Graph for Visual Inspection}

\author{Wei Jing$^{1,3,*}$, Di Deng$^{2,*}$, Zhe Xiao$^{3}$, Yong Liu$^{1,3}$ and Kenji Shimada$^{2}$ 
\thanks{$^{1}$ A*STAR Artificial Intelligence Initiative (A*AI); 1 Fusionopolis Way, 138632, Singapore. }%
\thanks{$^{2}$ Department of Mechanical Engineering, Carnegie Mellon University, 5000 Forbes Avenue, Pittsburgh, PA, 15213, USA. }%
\thanks{$^{3}$ Department of Computing Science, Institute of High Performance Computing; 1 Fusionopolis Way, 138632, Singapore. }%
\thanks{$^{*}$ The authors equally contribute to the work, email address: {\tt\small jing\_wei@ihpc.a-star.edu.sg, \tt\small dding@andrew.cmu.edu}}%
}

\begin{document}

\maketitle
\thispagestyle{empty}
\pagestyle{empty}

\begin{abstract}
 

Planning the path to gather the surface information of the target objects is crucial to improve the efficiency of and reduce the overall cost, for visual inspection applications with Unmanned Aerial Vehicles (UAVs). Coverage Path Planning (CPP) problem is often formulated for these inspection applications because of the coverage requirement. Traditionally, researchers usually plan and optimize the viewpoints to capture the surface information first, and then optimize the path to visit the selected viewpoints. In this paper, we propose a novel planning method to directly sample and plan the inspection path for a camera-equipped UAV to acquire visual and geometric information of the target structures as a video stream setting in complex 3D environment. The proposed planning method first generates via-points and path primitives around the target object by using sampling methods based on voxel dilation and subtraction. A novel Primitive Coverage Graph (PCG) is then proposed to encode the topological information, flying distances, and visibility information, with the sampled via-points and path primitives. Finally graph search is performed to find the resultant path in the PCG to complete the inspection task with the coverage requirements. The effectiveness of the proposed method is demonstrated through simulation and field tests in this paper.

\end{abstract}

\section{Introduction}


\noindent Recent years, vision-based structural inspection applications with Unmanned Aerial Vehicles (UAVs) in complex 3D environment have drawn increasingly attention from both industry and academia \cite{jing2016sampling} \cite{bircher2015structural} \cite{deng2017uav}. These inspection tasks require navigating a UAV around target objects to capture the visual information of the target surface area, for further investigation. The inspection is often a highly-repetitive task, and the target objects are in different sizes and geometries. Therefore, automatically generating efficient inspection paths for various target objects is crucial. Such planning task for inspection applications must satisfy surface coverage constraints, thus it is usually considered as a Coverage Path Planning (CPP) problem \cite{galceran2013survey}. 

For the path planning of UAV visual inspection applications, the 3D geometric model of a target is available, or it can be built prior to the planning process. The change in target structure and ambient environment is often neglectable and could be considered as static. Therefore, the planning problem for visual inspection is considered as a offline 3D coverage planning problem with model available. Given these application requirements, in this paper, we are interested in designing a offline, model-based coverage planning algorithm for structural inspection task with camera-equipped UAV.

\begin{figure}[!ht]
\centering
	\includegraphics[width=0.99\linewidth]{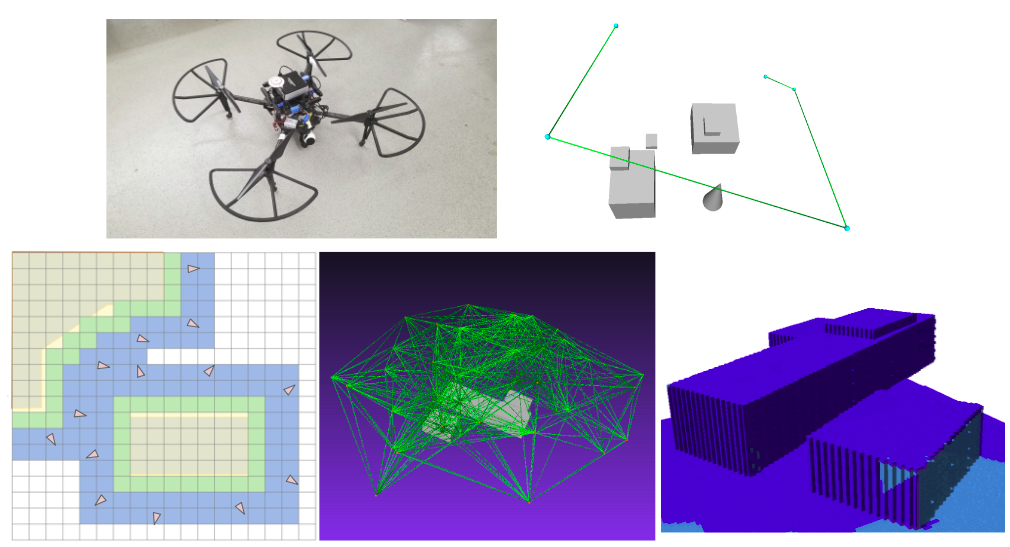}
	\caption{Path Planning for Structural Inspection with Camera-equipped UAV}\label{fig:frontpage}
\end{figure}

\begin{figure*}[!ht]
\centering
	\includegraphics[width=0.96\linewidth]{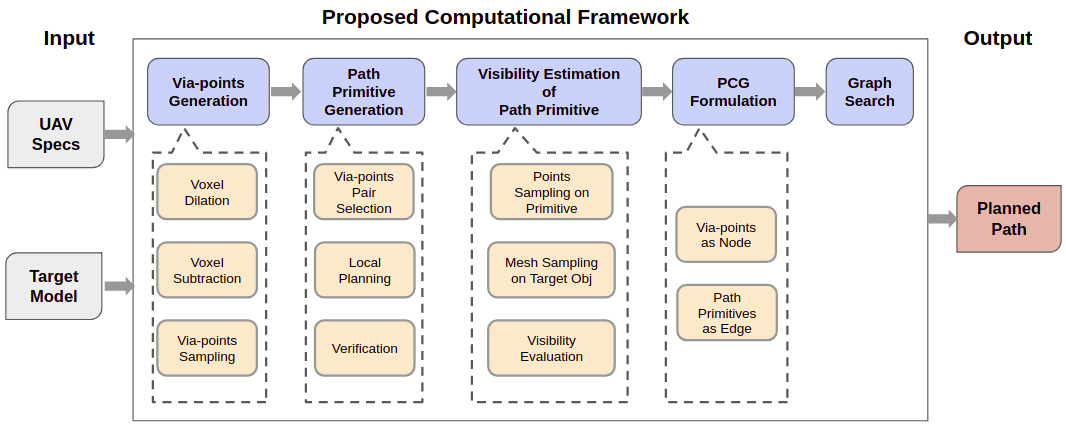}
	\caption{The Block Diagram of the Proposed Coverage Planning Framework}\label{fig:overview}
\end{figure*} 

In this paper, we propose a novel computational planning method for CPP problem with application to the structure inspection with UAV, where the UAV is capturing visual information while flying around the target structure. The proposed method first generates via-points and path primitives around the target building; then the topological information, together with the traveling cost and visibility information, are used to build the Primitive Coverage Graph (PCG); after that, graph search method is performed on the PCG to find a inspection path that satisfies the application requirements. The main contributions of this paper are:
\begin{itemize}
	\item a novel sampling-based coverage planning framework that is natively suitable for model-based, continuous CPP problem with UAV visual inspection, improved from our previous discrete viewpoint-based planning framework \cite{jing2016sampling} \cite{jing2016view}.
	\item flexible and reconfigurable design of the planning framework such that the modules (e.g. sampling method, local planner, visibility model) could be substituted with different algorithms in order to make the framework adaptable for requirements from different applications.
	\item a novel sampling-based path primitive and via-point generation method for structural inspection application with UAVs, which allows direct search on the paths rather than the discrete viewpoints; 
	\item a novel PCG-based strategy for the topology, coverage and cost information encoding and corresponding graph-search algorithms to solve the path finding-problem of the coverage planning.
\end{itemize}

\section{Relevant Work}

\noindent Research work has been conducted on the CPP problems intensively in past years \cite{bircher2015structural} \cite{englot2013three} \cite{scott2009model}, because many real-world applications could be formulated as CPP problems. 

Most of the previous work solves the CPP problem in two steps to address the requirement of visual information capturing. First, the planning algorithms identify a set of viewpoints that meet the coverage requirements for the inspection applications; the process of finding these viewpoints is called view planning or viewpoint planning problem (VPP) \cite{scott2003view} \cite{chen2011active}. Then the planning algorithms find the shortest path to maneuver a UAV/robot among the selected viewpoints to take the visual measurement, which is usually formulated as a (sequential) path planning problem \cite{alatartsev2015robotic} \cite{jing2017coverage}.

The sampling-based coverage planning method such as ``Generate-Test'' framework has been commonly applied to the model-based VPP in static, complex 3D environment, as summarized in the survey \cite{scott2003view}, where the methods sample a large set of candidate viewpoints, from which a subset of suitable viewpoints is selected according to the application requirements \cite{scott2009model} \cite{jing2016view} \cite{tarbox1995planning}. In addition, other view planning methods such as Next-Best-View (NBV) methods \cite{vasquez2014view} \cite{song2017online}, Reinforcement Learning based methods \cite{kaba2017reinforcement} and computational geometry based method \cite{scott2003view} have also been used to address different requirements in different VPP applications.

In addition to viewpoint planning, the UAV/robot path is also required to be planned for the inspection applications. Early work \cite{chen2004automatic} \cite{jing2017coverage} formulated the path planning problem as a Traveling Salesman Problem (TSP) after finding the required viewpoints. This often yields sub-optimal results due to the separation of the optimization processes. Later Coverage Planning with combining viewpoint and path planning problem into one-step optimization problem has also been proposed to achieve better optimization results \cite{jing2017redundant} \cite{wang2007view}. Additional previously planning methods for VPP and CPP problems in different applications could be found in several survey papers: \cite{galceran2013survey} \cite{chen2011active} \cite{almadhoun2016survey} \cite{scott2003view} \cite{choset2001coverage} \cite{tarabanis1995survey}.

However, most of these methods focused mainly on planning and optimization of discrete viewpoints for visual information capturing, and then find a optimal/sub-optimal path to visit the selected viewpoints \cite{chen2004automatic} \cite{jing2016view} \cite{roberts2017submodular}, or focused on coverage applications with continuous paths in simplified environment \cite{papadopoulos2013asymptotically} \cite{choset2001coverage}. Thus, these algorithms can not make full use of the video capturing features of the modern UAVs such that they are unsuitable or not performing well for our application. In this paper, instead of discrete viewpoints sampling and selection in the traditional sampling-based coverage planning methods \cite{scott2003view} \cite{jing2016sampling} \cite{jing2016view}, the proposed sampling-based coverage planning method directly samples the path primitives and then searches for a subset of the generated path primitive samples that satisfies the coverage requirements. 

\section{Proposed Method}

\noindent In this paper, we propose a novel planning framework for the CPP problem of building inspection applications with UAV, using voxel dilation and subtraction, path primitive sampling, and Primitive Coverage Graph (PCG) encoding and searching. The overall proposed framework can be summarized as followings: The first step is to generate the via-points around the target buildings; then the path primitives are generated based on these sampled via-points; after that, the third step is to estimate the visibility information of the path primitives; then a PCG is generated to encode the topological information, traveling cost, and visibility information; the last step is to solve the inspection path-finding problem through graph search. 

An overview of the proposed method is shown in Fig. \ref{fig:overview}. The computational framework is modularized and reconfigurable such that different algorithms could be plugin to address different requirements for various applications, for example, different sampling strategies, local planners, visibility models or search algorithms could be used in the corresponding modular to address the requirements from different applications.

\subsection{Via-points Sampling by Voxel Dilation and Subtraction}

\noindent In the proposed method, we first generate the via-points around the target structure in 3D space using the voxel-based sampling method, which utilizes importance sampling strategy to generate the via-points efficiently. This voxel-based via-points sampling method uses voxel dilation and subtraction to find the effective sampling region in the space. Our via-points generation method creates an initial feasible sampling region that satisfies both the safety and viewing-range requirements, scaling down the effective search of candidates for via-points, and accordingly facilitating the path primitive and PCG generation in later steps.

\begin{figure}[!ht]
\centering
	\includegraphics[width=0.7\linewidth]{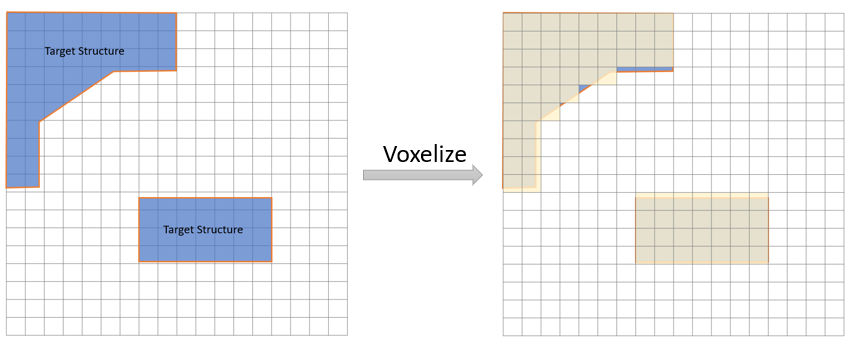}
	\caption{Voxelizing the polygonal model of the target structure}\label{fig:voxelize}
\end{figure}

\begin{figure}[!ht]
\centering
	\includegraphics[width=0.99\linewidth]{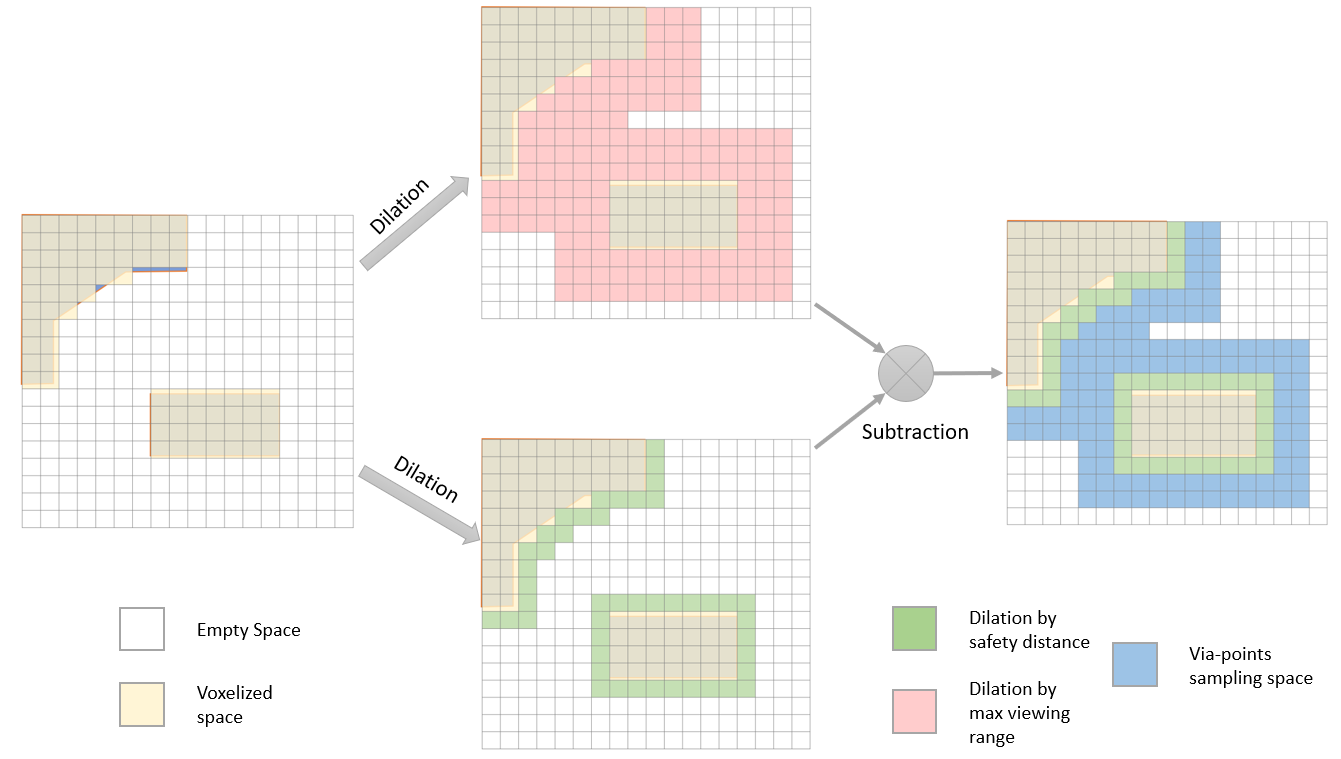}
	\caption{Generating sampling region for via-points: the original voxel model is dilated twice, once by the safety offset (shown in green) and once more by the max viewing range (shown in red); after that, a subtraction operation is performed, making the resultant region the sampling region (shown in blue)}\label{fig:vp_sample}
\end{figure} 

In the via-points generation step, the target structure is first voxelized to a 3D voxel model, as shown in Fig. \ref{fig:voxelize}. In order to compute efficient sampling region for the via-points, two voxel dilation operations are performed, one by the maximum viewing range and the other by a safety offset, as shown in Fig. \ref{fig:vp_sample}; after that, voxel subtraction operation is performed between the two dilated voxel model to get the valid sampling region in the 3D Euclidean space; sampling of via-points is then performed by randomly choosing voxels in the sampling region, and then randomizing a position within the voxel. The overall process of the via-point generation is shown in Fig. \ref{fig:voxelize} and \ref{fig:vp_sample}.

In addition to the position, the viewing direction $\bm{ \nu}$ of a given via-point located at $\mathbf{p_{vp}}$ is computed by the local potential field method \cite{jing2016sampling}, as shown in Eqn. \ref{eq::vp_dir}: 
\begin{align} \label{eq::vp_dir}
 &\bm{\nu} = \frac{\sum_i^N \frac{\mathbf{\mathbf{p_{vp}} - \mathbf{p_{patch_i}}}}{\Vert\mathbf{\mathbf{p_{vp}} - \mathbf{p_{patch_i}}}\Vert^3}}{\Vert\sum_i^N \frac{\mathbf{\mathbf{p_{vp}} - \mathbf{p_{patch_i}}}}{\Vert\mathbf{\mathbf{p_{vp}} - \mathbf{p_{patch_i}}}\Vert^3}\Vert}, \\
\text{for all: } \quad & \lbrace \mathbf{p_{patch_i}} \vert (\mathbf{p_{vp}} - \mathbf{p_{patch_i}})^T(\mathbf{p_{vp}} - \mathbf{p_{patch_i}}) \nonumber\\
 &\quad \quad \quad \quad < d_{max}\rbrace ,
\end{align} 
where $\mathbf{p_{patch_i}}$ is the position of the $i^{th}$ surface patch; $d_{max}$ is the maximum range of the local potential field.

\subsection{Path-Primitive Sampling}

\noindent Most previous ``Generate-Test" methods consider the sampled via-points as candidate viewpoints; then select a subset from these viewpoints to fulfill the coverage requirements of the application; and finally plan the inspection paths to visit the selected viewpoints to complete the inspection task \cite{chen2004automatic} \cite{scott2009model} \cite{jing2016view}. In this paper, however, we use these sampled points as via-points to generate path primitives, then directly select the path primitives as the final planning results. This approach is more intuitive and suitable to generate inspection path by considering the UAV inspection application as a continuous CPP problem in which a UAV continuously captures video stream whilst moving.

\begin{figure}[!ht]
\centering
	\includegraphics[width=0.8\linewidth]{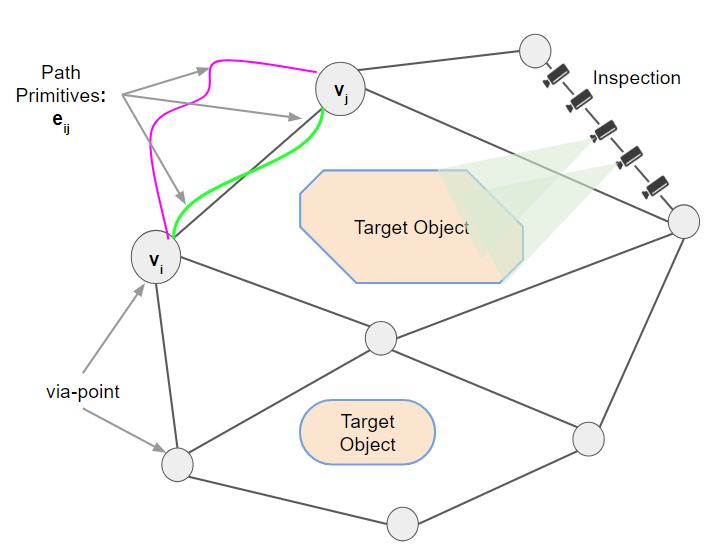}
	\caption{Simplified 2D illustration of the path primitives and estimation of the visibility information }\label{fig:illustration_2d}
\end{figure}

For the path primitive sampling process, we first choose the via-points pairs from the sampled via-points by randomly sampling pairs within a maximum distance; then a collision-free local planner $\mathcal{L}$ is used to find a local path that connects the via-point pair, as shown in Fig. \ref{fig:illustration_2d}. In this paper, we directly connect two via-points with straight line as local planning process, with the collision check of the target objects. Other sampling-based planners such as Rapidly exploring Random Tree (RRT) \cite{lavalle2006planning} and Probabilistic Roadmap (PRM) \cite{kavraki1996probabilistic} and their variations could also be used as local planner, for inspection applications such as robotic arms in more confined space to achieve better local planning performance.

An example of the sampled path primitives is illustrated in Fig. \ref{fig:traj1}. The viewing direction of any point $\mathbf{p_{t}}$ on the path primitives is computed by the interpolation of the viewing directions at the starting and ending via-points to ensure smooth transitions of angles while the UAV is moving, as shown in Eqn. \ref{eq::traj_dir}.
\begin{align} \label{eq::traj_dir}
 &\bm{\nu}_t = \bm{\nu}_s + \frac{ \Vert \mathbf{p_{t}} - \mathbf{p_{s}}\Vert }{\Vert \mathbf{p_{e}} - \mathbf{p_{s}}\Vert } (\bm{\nu}_e - \bm{\nu}_s) 
\end{align} 
where $\bm{\nu}_t$ and $\mathbf{p_{t}}$ are the viewing direction and position at any point on the path primitive; $\bm{\nu}_s,  \mathbf{p_{s}}, \bm{\nu}_e $ and $\mathbf{p_{e}}$ are the viewing directions and positions at the starting and ending via-points, respectively.

Overall, given the planning parameters, the algorithm to sample the path primitives and the via-points is shown in Algo \ref{algo:primitive}.
\begin{algorithm}
\small
\caption{Path Primitive Sampling Algorithm } 
\label{algo:primitive}
\begin{algorithmic}[1]
    \Require
      		The max distance for primitive sampling, $d_{max}$; the max viewing range of sensor, $d_{vis}$; the safety distance, $d_{safe}$; the collision-free Local Planner, $\mathcal{L}$, the 3D model of target objects $ \mathbf{M}$; 
    \Ensure
     		The set of sampled via-points $\mathbf{V}$; the set of sampled path primitives, $\mathbf{E}$; 
\State $ \mathbf{E}, \mathbf{V} \leftarrow \emptyset, \emptyset$
\State $D_1 \leftarrow $ dilate($M, d_{vis}$)
\State $D_2 \leftarrow $ dilate($M, d_{safe}$)
\State $D \leftarrow $ VoxelSubstract($D_1, D_2$)
\While{sizeof$(\mathbf{V}) < n_{vp}$}
	\State $v \leftarrow$ RandomSampleVP($D$)	
	\State $\mathbf{V} \leftarrow$ append($\mathbf{V}, v$)
\EndWhile
\For{$v_i, v_j \in \mathbf{V}, i \neq j$}
\If{ dist($v_i, v_j$) $ \leq d_{max}$ }
\State $e_{ij} \leftarrow \mathcal{L}(v_i, v_j, M)$ 
\State $\mathbf{E} \leftarrow $  Append($ \mathbf{E}, e_{ij}$)
\EndIf
\EndFor
\State \Return{$\mathbf{V}, \mathbf{E}$}
\end{algorithmic}
\end{algorithm}

\subsection{Visibility Estimation of Path Primitives} \label{sec::vis}


\noindent The visibility information is modeled as binary information that is pre-computed for each sampled path primitive, to avoid repetitive computation during the optimization process. Previous viewpoints-based approaches also pre-compute visibility, but only for individual viewpoint rather than path \cite{scott2009model} \cite{jing2016sampling}.

\begin{figure}[!ht]
\centering
	\includegraphics[width=0.8\linewidth]{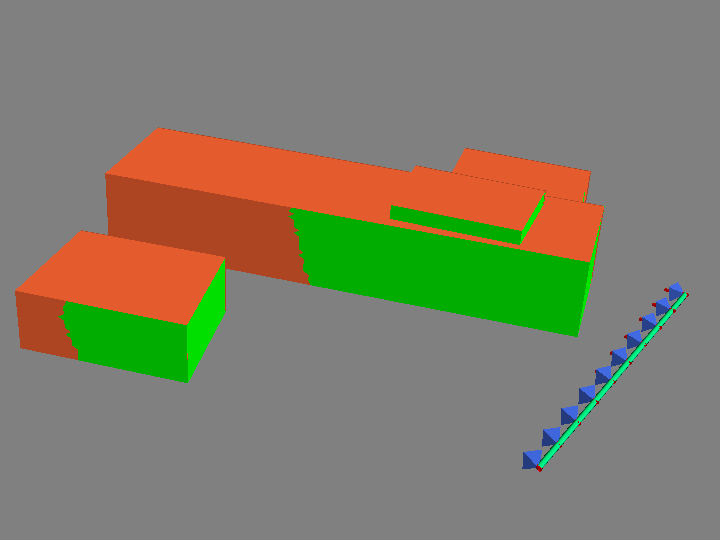}
	\caption{Visibility estimation of path primitive, green areas are visible to the path primitive shown in the figure, computed with the visibility model. The viewpoints sampled on the path primitive are used to estimate the visibility, as shown in the figure }\label{fig:illustration_3d}
\end{figure} 


In this paper, we propose a sampling-based method to estimate the visibility of the path primitive and the target building. The proposed visibility estimation method first samples viewpoints from the path primitives and estimate the binary visibility information of the surface patches and the sampled viewpoints, as shown in Fig. \ref{fig:illustration_2d} and Fig. \ref{fig:illustration_3d}. The surface of the target object is subdivided to small and uniform triangular patches using the Bubble Mesh method \cite{shimada1995bubble}, for the visibility evaluation. After estimating the visibility of the sampled viewpoints, we combine the visibility information from all sampled viewpoints on a path primitive to compute the overall visibility information of the path primitive. 

The visibility evaluation model between a individual sampled viewpoint and a surface patch is drawn from our previous work \cite{jing2016view}\cite{jing2016sampling}\cite{deng2017uav}, and similar to \cite{scott2009model}\cite{scott2003view}, as listed below:
\begin{itemize}
	\item The surface patch must be in the Field of View of the UAV from the sampled viewpoint.
	\item The surface patch must be within the viewing range of the UAV from the sampled viewpoint.
	\item The viewing angle must be within a certain range predicted by the sensor specifications.  
	\item There must be no occlusion between the sampled viewpoint and surface patch.
\end{itemize}

 Then the visibility information for a path primitive is combined from the sampled viewpoints and represented in a $m \times 1$ binary vector $\mathbf{s}$, where $m$ is the number of surface patches of the target structure. A 3D visualization results of visibility estimation of a given path primitive is shown in Fig. \ref{fig:illustration_3d}. It will be associated with the corresponding edge and stored in the Primitive Coverage Graph (PCG).

\subsection{Primitive Coverage Graph for Information Encoding} \label{sec:pcg}

\begin{figure}[!ht]
\centering
	\includegraphics[width=0.99\linewidth]{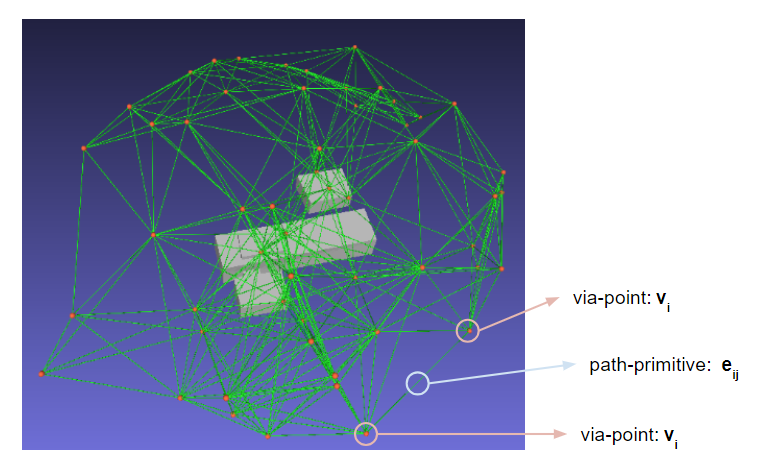}
	\caption{Illustration of sampled path primitives and via-points in 3D space (this figure is for visualization, the via-points and path primitives are much denser in the implementation)}\label{fig:traj1}
\end{figure} 

\noindent In this paper, we propose a novel Primitive Coverage Graph (PCG) representation $\mathcal{G}(\mathbf{V}, \mathbf{E})$ to encode the topological information, UAV flying distance (traveling cost) as well as visibility information of the sampled via-points and path primitives. As visualized in Fig. \ref{fig:traj1}, node $v_i \in \mathbf{V}$ is a via-point and edge $e_{ij} \in \mathbf{E} $ represents a path primitive that connects via-points $v_i$ and $v_j$. 

For the PCG proposed in this paper, the key information is associated with the edges $\mathbf{E}$ that correspond to the sampled path primitives. The nodes $\mathbf{V}$ encode the topological information between the path primitives, for the assurance of the connectivity and no-branching of the final path in the graph search in later step. The information associated with edge $e_{ij}$ is $\lbrace d_{ij}, \mathbf{s}_{ij} \rbrace $, where $d_{ij}$ is the length of the path primitive; $\mathbf{s}_{ij}$ is a $m \times 1$ vector to model the binary visibility information of sampled patches of the target structure. The visibility information of a path primitive is computed in Sect. \ref{sec::vis}. For element $s_{ij}^k \in \mathbf{s}_{ij}, k\in [1, m]$, $s_{ij}^k = 1$ indicates a surface patch indexed by $k$ is visible to path primitive $e_{ij}$, and $s_{ij}^k = 0$ indicates it is invisible. The PCG representation and information encoding not only ensures the graph search step is able to find a fully connected path without branches, but also ensures that the coverage constraints are satisfied.

The PCG construction methods are summarized in Algo. \ref{algo:pcg}. It takes the sampled via-points $\mathbf{V}$ and the sampled path primitives $\mathbf{E}$ from Algo. \ref{algo:primitive} as input, and generate the PCG $\mathcal{G}(\mathbf{V}, \mathbf{E})$ as output to encode the information. 

\begin{algorithm}
\small
\caption{Primitive Coverage Graph Construction Algorithm}
\label{algo:pcg}
\begin{algorithmic}[1]
    \Require
      		The 3D model of target objects, $ \mathbf{M}$, the set of sampled via-points, $\mathbf{V}$; the set of sampled path primitives, $\mathbf{E}$;
    \Ensure
     		The Primitive Coverage Graph, $\mathcal{G}(\mathbf{V}, \mathbf{E})$; 
\For{$e_{ij} \in \mathbf{E}$}
\State $\mathbf{s}_{ij} \leftarrow $ VisibilityEstimation($e_{ij}, \mathbf{M}$)
\State ${d}_{ij} \leftarrow $ ComputeDist($e_{ij}$)
\State ${e}_{ij} \leftarrow $ Associate($e_{ij}, \mathbf{s}_{ij}, {d}_{ij}$)
\EndFor
\State $ \mathcal{G}(\mathbf{V}, \mathbf{E}) \leftarrow $ BuildGraph($\mathbf{V}, \mathbf{E}$) 
\State \Return{$\mathcal{G}(\mathbf{V}, \mathbf{E})$}
\end{algorithmic}
\end{algorithm}
\noindent where \textit{VisibilityEstimation} described in Sect. \ref{sec::vis} is to estimate the visibility information of each sampled path primitive.



\subsection{Primitive Coverage Graph Search}



\noindent The path finding problem is formulated with the PCG, $\mathcal{G}(\mathbf{V}, \mathbf{E})$, that encodes the topological information, traveling distances and coverage information of the path primitives. The objective of the path planning problem is to find a fully connected, branch-less path from the PCG that minimize the traveling distance while the required coverage ratio $\delta_d$ is achieved.  

In this paper, we propose a Greedy Neighborhood Search (GNS) algorithm to search the final path directly for the CPP problem, with all the information encoded in the PCG. The details of the algorithm is shown in Algo. \ref{algo:greedy}.
\begin{algorithm}
\small
\caption{Greedy Neighborhood Search Algorithm for Path Finding on PCG}
\label{algo:greedy}
\begin{algorithmic}[1]
    \Require
      		The desired coverage ratio, $\delta_d$; the Primitive Coverage Graph, $\mathcal{G}(\mathbf{V}, \mathbf{E})$, and other algorithm specific parameters; 
    \Ensure
     		  the resultant path primitives $ \mathbf{E}_{res}$; 
\State $ \mathbf{V}_{res}, \mathbf{E}_{res}, \delta \leftarrow \emptyset, \emptyset, 0$
\State $\delta \leftarrow $ FindCoverage$(\mathcal{G}, e)$
\State $ v, e \leftarrow $ Initialize$(\mathbf{\mathcal{G}})$
\While{$\delta \leq \delta_d$}
\State $ \mathbf{E}_{nbr} \leftarrow$ FindNeighbour($v, \mathcal{G}$)
\State $v, e \leftarrow$ FindTraj($\mathbf{E}_{nbr}, \mathbf{E}_{res}, \mathcal{G}$)
\State $\mathbf{V}_{res} \leftarrow$  Append($\mathbf{V}_{res}, v$) 
\State $\mathbf{E}_{res} \leftarrow$  Append($\mathbf{E}_{res}, e$) 
\State $\delta \leftarrow $ FindCoverage$(\mathcal{G}, \mathbf{E}_{res})$
\EndWhile
\State \Return{$\mathbf{E}_{res}$}
\end{algorithmic}
\end{algorithm}

For the proposed GNS algorithm, at each iteration in the loop, the algorithm searches the current neighborhood of the path primitives that yields highest incremental coverage over distance. The path primitive will be added to the solution set $\mathbf{E}_{res}$ that consists of a number of path primitives with order. Then the coverage information and neighbourhood will be updated. The iteration will repeat until the desired coverage ratio, $\delta_d$, is achieved. 

The \textit{FindTraj} in GNS is to find the path primitive in the neighbourhood, based on the incremental coverage over distance, given the current via-point and path primitives, as shown in Eqn. \ref{eq:best_traj}.
\begin{equation} \label{eq:best_traj}
e = \argmax_{e \in \mathbf{E}_{nbr}} \frac{\triangle cov(e, \mathbf{E}_{res})}{d_e} 
\end{equation}
where $d_e$ is the flying distance of the path primitive associated with edge $e$; $\mathbf{E}_{nbr}$ is the neighbourhood edges (the edges connected to current via-point $v$), identified by \textit{FindNeighbour} using the topological information in $\mathcal{G}$; $\mathbf{E}_{res}$ is the current set of edges in the resultant path, which provides the total current coverage information; $\triangle cov(e, \mathbf{E}_{res})$ is the incremental surface coverage by the path primitive, $e$. 

\section{Results}

\subsection{Setup}

\noindent In this paper, simulation and experimental tests are conducted to validate the effectiveness of the proposed planning method. In the simulation tests, we first applied the proposed planning method to compute the inspection paths; then the drone is flied on the planned paths in Drake \cite{drake}, a toolbox for robotic simulation and analysis, to capture the visual information; after that the Octomap library \cite{hornung13auro} is used to build the 3D voxel model of the target structure to validate inspection results of the computed UAV paths. In the real-world experiment, a DJI drone is used to capture the required surface information with the planned paths, and the 3D surface of the target objects is reproduced to further demonstrate the effectiveness of the proposed method.

\subsection{Simulation Tests with Drake and Octomap}

\subsubsection{Planning Parameters and Results}

\begin{figure}[!ht]
\centering
\begin{subfigure}[b]{0.99\linewidth}
	\includegraphics[width=0.92\linewidth]{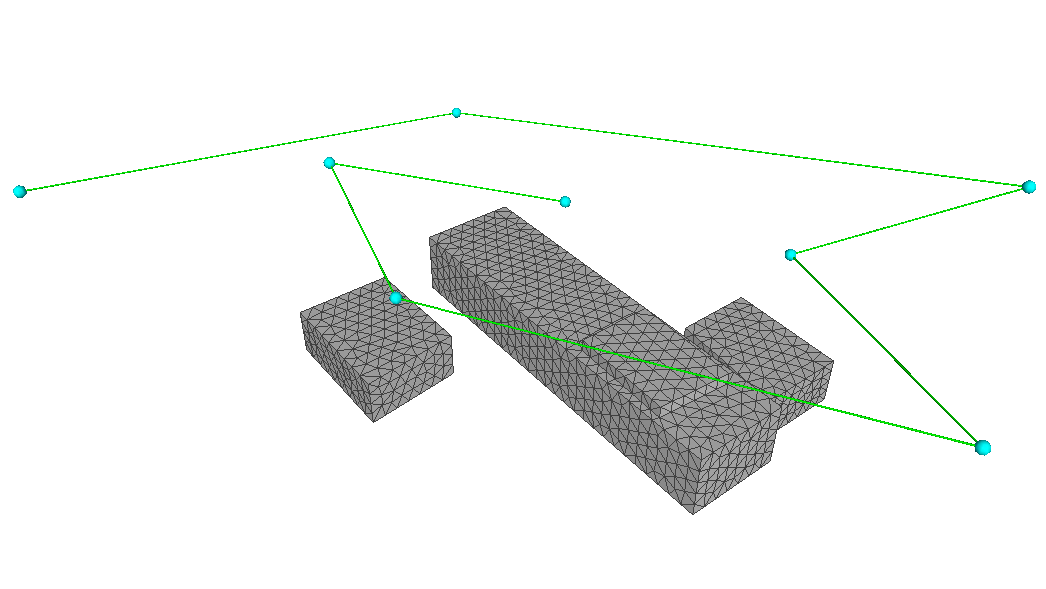}
	\caption{Target Structure 1}\label{fig:planned_path_ny7}
\end{subfigure}    

\begin{subfigure}[b]{0.99\linewidth}
	\includegraphics[width=0.92\linewidth]{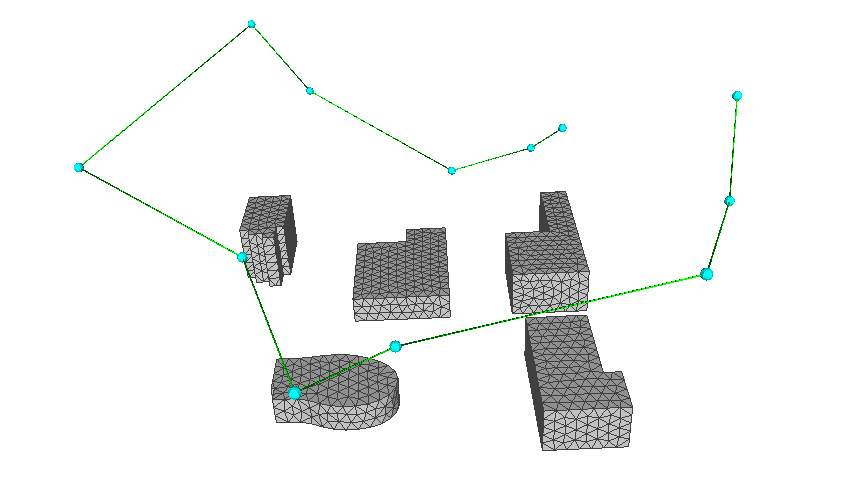}
	\caption{Target Structure 2}\label{fig:planned_path_b42}
\end{subfigure} 
\caption{Visualization of planned UAV inspection path around the target buildings}
\end{figure}

First, the proposed method is used to compute the inspection path of the UAV for the two different target buildings. $1,765$ and $1,410$ path primitives are generated around each target structure, respectively. The parameters used for the coverage planning process is shown in Table \ref{table::cpp_parameters_1}. The resultant paths computed by the proposed method are visualized in Figs. \ref{fig:planned_path_ny7} and \ref{fig:planned_path_b42}.
\begin{table}[!ht]
\small
	\centering
	\caption{Planning parameters used in this paper} \label{table::cpp_parameters_1}
\begin{tabular}{@{}cc@{}}
\toprule
\textbf{Parameter}      & \textbf{Value} \\ \midrule
FOV (Diagonal)          & $94  \degree  $         \\
Max Viewing Angle       &   $ 75 \degree$             \\
Resolution (Pixels)     & $4000 \times 3000 $   \\
Coverage Ratio ($\delta_d$)        & $99.0\%   $         \\
Safety Distance         & $2m   $         \\
Max Viewing Range       & $50m   $         \\ \bottomrule
\end{tabular}
\end{table}

The result comparison with two previous methods is shown in Table \ref{table::benchmark}, based on the average of 10 trials. The flying distances computed by the proposed method in this paper are 425.6m and 466.2m for the two target structures, respectively. The distances of the inspection paths are 507.7m and 587.5m using the previous methods \cite{jing2016view}\cite{jing2017coverage} that solves the TSP after applying the viewpoint-based methods to find the viewpoints (labeled as VPP-TSP). The discrete viewpoints based greedy method, a simple but robust baseline \cite{kaba2017reinforcement}\cite{jing2017coverage} that iteratively adds the most cost effective viewpoints, planned the path with the traveling distance of 531.0m and 687.1m. Assuming a constant flying speed, the reductions of the required inspection time are 18.4\% and 29.2\% on average, compared to the two previous methods, respectively.

\begin{table}[!ht]
\small
	\centering
	\caption{Planning results} \label{table::benchmark}
\begin{tabular}{@{}ccc@{}}
\toprule
\textbf{Methods}                                    & \textbf{Structure 1}  & \textbf{Structure 2}\\ \midrule
The Proposed Method                                 & $425.6m$              & $466.2m$       \\ 
VPP-TSP\cite{jing2016view}\cite{jing2017coverage}   & $ 507.7m$             &  $587.5m$         \\
Greedy Method                                       & $ 531.0m   $          &     $687.1 m   $        \\ \bottomrule
\end{tabular}
\end{table}

\subsubsection{Computational Simulation Tests}

\begin{figure}[!ht]
\centering
	\includegraphics[width=1\linewidth]{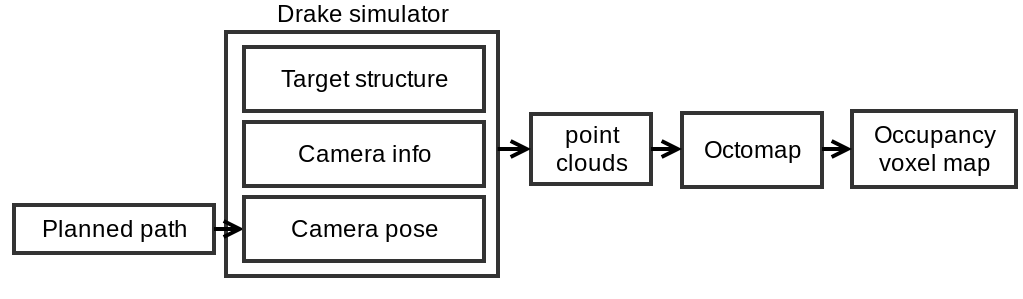}
	\caption{Simulation Flowchart with Drake and Octomap} \label{fig:drake_simulation}
\end{figure}

\begin{figure}[!ht]
\centering
\begin{subfigure}[b]{0.99\linewidth}
	\includegraphics[width=0.92\linewidth]{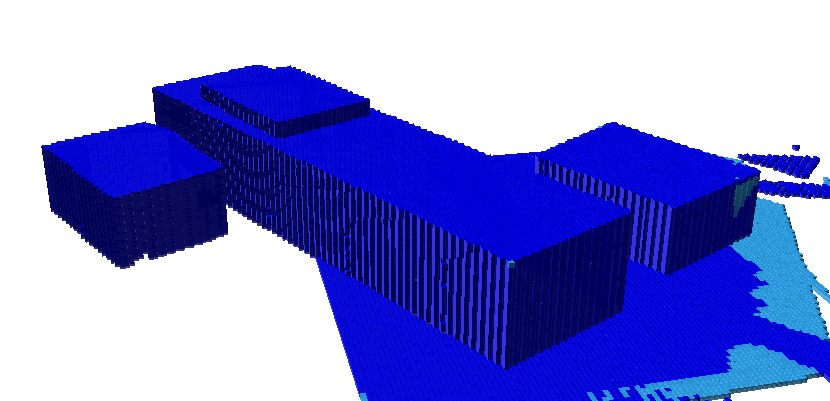}
	\caption{Target Structure 1}\label{fig:octo_ny7}
\end{subfigure}    

\begin{subfigure}[b]{0.99\linewidth}
	\includegraphics[width=0.92\linewidth]{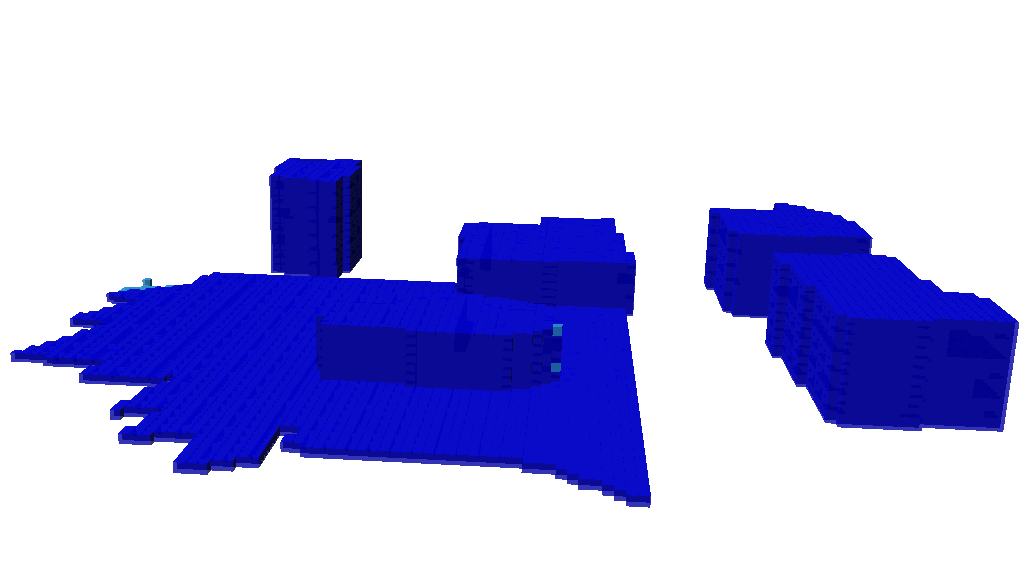}
	\caption{Target Structure 2}\label{fig:octo_b42}
\end{subfigure} 
	\caption{Voxel model of results from Drake-Octomap simulation using the planned paths with the proposed algorithm}\label{fig:res_ny_ros}
\end{figure}

Simulation tests implemented with Drake simulator \cite{drake} are performed to validate the coverage and effectiveness of the proposed method. Fig. \ref{fig:drake_simulation} shows the simulation test work-flow. With the camera simulator provided by Drake, point cloud data is generated based on the camera poses. The generated point cloud data in each frame is located based on the world frame through Eqn. \ref{eq:3d_tf}:
\begin{equation} \label{eq:3d_tf}
 \mathbf{P}^{world} = \mathbf{T}^{world}_{cam} \mathbf{P}^{cam},   
\end{equation}
where $\mathbf{P}^{world}$ is the position of the point cloud data in the world frame, $\mathbf{T}^{world}_{cam}$ is the pose of the camera in the world frame and $ \mathbf{P}^{cam}$ is the position of the point cloud with respect to the camera frame. The Octomap library \cite{hornung13auro} is then used to create the occupancy voxel models from the point cloud results. The visualization of the reconstructed voxel models is shown in Fig. \ref{fig:res_ny_ros}.

\subsection{Field Test}

\begin{figure}[!htbp]
\centering
	\includegraphics[width=0.75\linewidth]{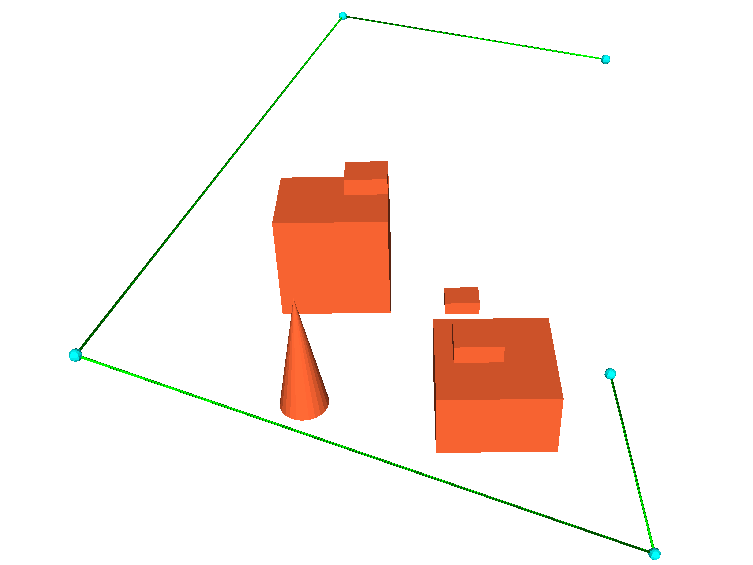}
	\caption{Visualization of the Planned Path for the Experiment}\label{fig:planned_path_exp}
\end{figure}

\noindent An field test has also been conducted to further validate the proposed planning method in real-world scenario. In the field test, a DJI Matrice 100 is used to fly around the target structure to capture the surface information, with the path computed by the proposed method. The max viewing range used for this experiment is 10m, in order to get better details for reconstructions for in practical applications; the safety distance used for the experimental test is 0.5m. The planned path is shown in Fig. \ref{fig:planned_path_exp}. 

\begin{figure}[!htbp]
\centering
	\includegraphics[width=0.95\linewidth]{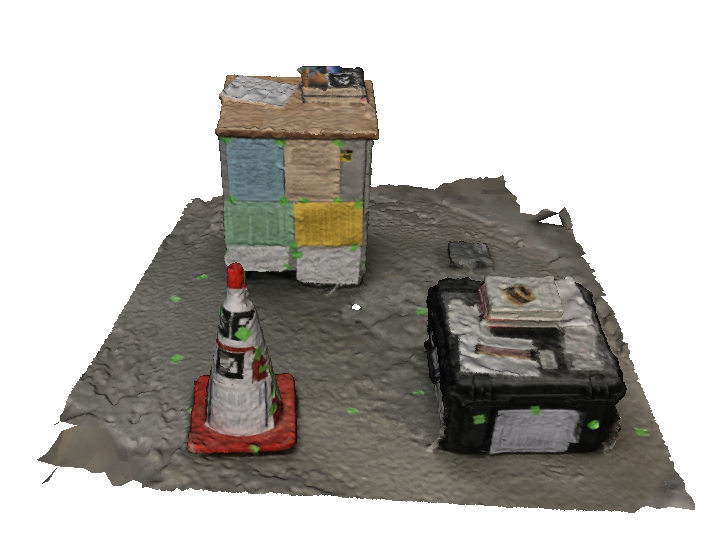}
	\caption{Visualization of 3D reconstruction results of the field tests}\label{fig:field_exp}
\end{figure}

\begin{figure}[!htbp]
\centering
	\includegraphics[width=0.99\linewidth]{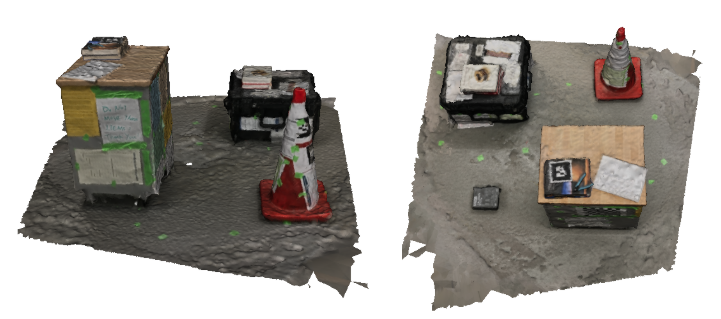}
	\caption{Additional views of 3D reconstruction results of the field tests}\label{fig:field_exp2}
\end{figure}

The 3D visualization of the resultant reconstructed model is shown in Figs. \ref{fig:field_exp} and \ref{fig:field_exp2}. The reconstruction is done by using Structure From Motion (SFM) method with 87 pictures sampled from the video taken by the drone, during the inspection process. VisualSFM \cite{wu2011visualsfm} and CMP-MVS \cite{jancosek2011multi} are used in this papaer for SFM and surface reconstruction.

\section{Discussion}

\noindent The proposed planning method is shown to be able to automatic generate efficient path for different target objects in 3D environment, with the coverage requirements. Both simulation and field tests validate that the proposed planning method successfully covered the surface area of the target objects. 

Compared to previous discrete viewpoints sampling-based methods \cite{jing2016sampling} \cite{scott2009model}, the proposed path primitive sampling methods is able to find shorter path for the UAV inspection tasks, which helps to improve the efficiency and reduces the cost. The proposed method targets for the modern UAV system with high performance of continuous video capturing features, which is more naturally suitable for these applications, while the viewpoints sampling based method requires the UAV to take pictures at discrete viewpoints. Moreover, the discrete viewpoint sampling method is known to be unreliable for registration \cite{jing2017coverage}, because the viewing angles from different viewpoints may cause failure in image registration even if there are enough overlap areas of two images. For path primitive sampling method, however, the registration problem is usually not a concern because the images are sampled from continuous motion.

Additionally, the proposed planning framework is flexible and reconfigurable. Different algorithms/modules could be used to replace the current ones such that the proposed framework is adoptable for other applications, for example, product shape inspection problem with robotic manipulator \cite{jing2017modelbasedplanning} or mobile robot \cite{vasquez2014view}. Moreover, with properly modeling of the process, it is also possible to extend the proposed planning framework to other planning problem in robotic applications with coverage constraints, such as robotized painting, surface polishing.

\section{Conclusion}

\noindent In this paper, we presented a novel path-primitive based, offline planning framework for the continuous CPP problem, with applications of surface inspection using UAV. The proposed method utilizes voxel-based via-points sampling, path primitive sampling, PCG encoding, and graph search to plan UAV inspection paths that satisfy the coverage requirements. We also demonstrated the effectiveness of the proposed framework and the implementation in both simulation and experimental tests with different target objects. Through the comparison with previous methods, the reductions on the required inspection time are found to be 18.4\% and 29.2\% on average. 


\section*{Acknowledgement}

\noindent This research is supported by the Agency for Science, Technology and Research (A*STAR), Singapore, under its AME Programmatic Funding Scheme (Project \#A18A2b0046).

\bibliography{vpuav18}

\end{document}